\documentclass{article}
\usepackage{spconf,amsmath,graphicx}

\usepackage{algcompatible}

\usepackage{algorithm}
\usepackage{url}
\usepackage{algpseudocode}

\usepackage[english]{babel}

\title{Approximate Sampling using an Accelerated Metropolis-Hastings based on Bayesian Optimization and Gaussian Processes}
\twoauthors
  {Asif J. Chowdhury}
	{University of South Carolina\\
	Computer Science and Engineering\\
	Columbia, SC 29201, USA}
  {Gabriel Terejanu}
	{University of North Carolina Charlotte\\
	Department of Computer Science\\
	Charlotte, NC 28262, USA}
\begin{document}
%
\maketitle
\begin{abstract}
Markov Chain Monte Carlo (MCMC) methods have a drawback when working with a target distribution or likelihood function that is computationally expensive to evaluate, specially when working with big data. This paper focuses on Metropolis-Hastings (MH) algorithm for unimodal distributions. Here, an enhanced MH algorithm is proposed that requires less number of expensive function evaluations, has shorter burn-in period, and uses a better proposal distribution. The main innovations include the use of Bayesian optimization to reach the high probability region quickly, emulating the target distribution using Gaussian processes (GP), and using Laplace approximation of the GP to build a proposal distribution that captures the underlying correlation better. The experiments show significant improvement over the regular MH. Statistical comparison between the results from two algorithms is presented.
\end{abstract}
\section{Introduction}
\label{sec:intro}

Markov Chain Monte Carlo (MCMC) algorithms are widely applied in numerous fields of science, engineering and statistics ~\cite{a1mcmc} to sample from a target probability distribution. In this paper we focus on the Metropolis-Hastings (MH) algorithm, one of the premier algorithms under this class. MH, typically runs for a large number of iterations where, for each iteration, a new point on the parameter space is proposed using a proposal distribution; and the target function needs to be evaluated at the proposed point, based on which the algorithm decides whether to move to the proposed point or to stay in the same location. Thus, the target function needs to be evaluated many times. Reducing the number of steps is not an option as that diminishes the quality of the samples. For many physics based models, the forward simulations are expensive and performing these for each iteration incur large computational cost and becomes a performance bottleneck. Moreover, the initial samples of the Markov chain usually follow a distribution that is different from the target, and have to be discarded until the convergence to the target - termed as burn-in period - which is wasteful. In this work, we propose an enhanced MH algorithm, named MHGP that addresses these problems.

Different approaches have been proposed to make MCMC faster - using parallelization through multithreading ~\cite{prevwork2}; distributed algorithms to achieve better performance ~\cite{prevwork3}; reducing the computational cost of the accept/reject step by using smaller fraction of data ~\cite{prevwork1} or using log-likelihood estimator to work with random subset of the observations~\cite{prevwork5}. These methods, while, trying to reduce the cost of the target evaluation, do not actually reduce the number of times the target is evaluated. Gaussian process model has been proposed to off-load some of the computational work in Hybrid Monte Carlo ~\cite{gphmc}; approximation methods have been proposed ~\cite{simwork} where acceptance probabilities are calculated on a local approximation and the actual target is only evaluated once the proposal has been accepted. However, it would still require large number of target evaluations once the chain reaches high density region as more of the proposals are accepted there. Gaussian process approximation of the target distribution was used by the authors to decrease expensive function calls ~\cite{ownwork}. But, as more numbers are added to the GP, the computational cost increases as the time complexity for GP is $O(N^3)$. To improve the proposal distribution in MCMC, an adaptive approach has been used ~\cite{Haario99adaptiveproposal,mcmcdram} where information from simulation is utilized to adapt the proposal distribution. In Ref.~~\cite{Larjo2015}, a multi-step proposal distribution was introduced to speed up convergence by adjusting the proposal. The adaptive approach has the limitation that for high dimensional space the stationary distribution tends to be biased, which reduces the domain of the set of problems where this approach can be applied.

In this paper we propose an enhanced MH method. It uses Bayesian optimization to speed up the burn-in process and quickly reach high density region. Next, continuing with the GP obtained from the Bayesian optimization, Laplace approximation of the GP is taken around the peak to get the covariance matrix for an informed proposal distribution; and then, guided by this proposal, sample-generating iterations run as more training points are added to the GP, which continues to gain better approximation of the unnormalized target distribution. Due to positivity of probability density functions (pdf), GP is used to approximate the log of the target pdf instead of using it directly to the original pdf. The uncertainty measure of the GP predictions provides the uncertainty for the acceptance rate, which is then used to decide whether the objective function needs to be evaluated or it can be read from the GP approximation - resulting in fewer forward simulations as the iterations progress. Local Gaussian process~\cite{NIPS2014_5594,lgp} was used to avoid expensive calculations involving cumulative sampled points. The proposed algorithm was evaluated for different benchmark problems, two of which are presented here. The obtained samples are compared with those from plain MH and DRAM methods which reveal that samples from MHGP have no statistically significant difference from the established methods but is able to achieve similar target approximation with far fewer target evaluations.

\section{Methodology}

MHGP starts with initiation of a Gaussian process for Bayesian Optimization, which is a sequential approach to optimize an objective function by balancing between exploitation and exploration, controlled by an acquisition function ~\cite{Shahriari:2015}. It enables MHGP to reach the high probability region of the function in a handful of iterations. The optimized point and the GP provided by Bayesian optimization is used in the next step to come up with an informed proposal distribution that captures and approximate shape of the target distribution. We calculate Hessian of the GP at the mode of the distribution to apply Laplace approximation and thus obtain a multivariate Gaussian distribution with mean at the optimized point and a covariance matrix that we will use as the covariance of our proposal distribution for the following stages. 

As Bayesian Optimization often requires only a few steps to reach the optimized region, the Gaussian process may not be good enough to approximate the target distribution at the end of the first phase and the covariance of the proposal may not be positive semi-definite. In order to get a better approximation, a random walk, governed by some isotropic Gaussian proposal, is initiated starting from the optimized point obtained in the first phase of the algorithm. New points are added to GP by evaluating the objective function. The number of steps this needs to go on can be pre-specified or can be adaptively controlled by checking the uncertainty in the GP estimation on subsequent steps.

Next, MHGP enters its sample generating iterations. At each iteration, a new point is proposed centering at the current point with a covariance obtained from the previous phase. If the GP prediction at the proposed point has high uncertainty, the target needs to be evaluated and added to the GP training set, otherwise it is provided by the GP approximation. A high-level pseudo-code of MHGP is presented here:

\begin{algorithm}[H]

\caption{Proposed MHGP Method}

\label{MHGP}

\begin{algorithmic}[1]

\Procedure{MHGP}{}

\State Initialize $x^{(0)}$

\State Run \textit{BayesOpt} starting from $x^{(0)}$; it returns \textit{GP}

\State Set proposal distribution $Q$ to be the covariance obtained from Laplace approximation of \textit{GP}

\State Set $x^{(0)}$ to be the optimized point

\For{\textit{i} = 0 to \textit{N} - 1}

\State Sample $ u \sim\ U_{[0,1]} $

\State Sample $ x^* \sim\ Q(x^*|x^{(i)}) $

\State $ \mu{},\Sigma{} = \Call{getTargetValue}{x^*,x^{(i)}} $

\State $\textit{acceptanceRatio} = e^{\mu{}^*-\mu{}+\frac{\sigma{}_{xx}^2 + \sigma{}_{x^*x^*}^2 - 2\sigma{}_{xx^*}^2}{2}}$

\State $A(x^{(i)},x^*) = min \left\{ 1, \textit{acceptanceRatio} \right\}$

\If{ $u < A(x^{(i)},x^*)$ }

\State $ x^{(i+1)} = x^* $

\Else

\State $ x^{(i+1)} = x^{(i)} $

\EndIf

\EndFor

\EndProcedure

\end{algorithmic}

\end{algorithm}

\begin{algorithm}[H]

\caption{ Get Predicted or Evaluated Target Value }

\label{MHGP_train}

\begin{algorithmic}[1]

\Procedure{getTargetValue}{$x^*,x^{(i)}$}

\State $ \mu{},\Sigma{} = LocalGP(predict(x^*,x^{(i)})) $

\If{ $ \sqrt{e^{\sigma{}_{xx}^2 + \sigma{}_{x^*x^*}^2 - 2\sigma{}_{xx^*}^2} - 1} > threshold $ }

\If{ $p(x)$ was obtained by evaluating target}

\State Evaluate $ p(x^*) $; add to GP training set

\Else

\State Evaluate $p(x)$; add to GP training set

\If{ ratio is still greater than threshold }

\State Evaluate $p(x^*)$; add to GP training set

\Else

\State get $p(x^*)$ from LocalGP's prediction

\EndIf

\EndIf

\Else

\State get $p(x^*)$ from LocalGP's prediction

\EndIf

\State Return predicted or evaluated $p(x^*)$

\EndProcedure

\end{algorithmic}

\end{algorithm}

\begin{figure}[thb]

\begin{minipage}[b]{1\linewidth}
  \centering
  \centerline{\includegraphics[width=9.0cm]{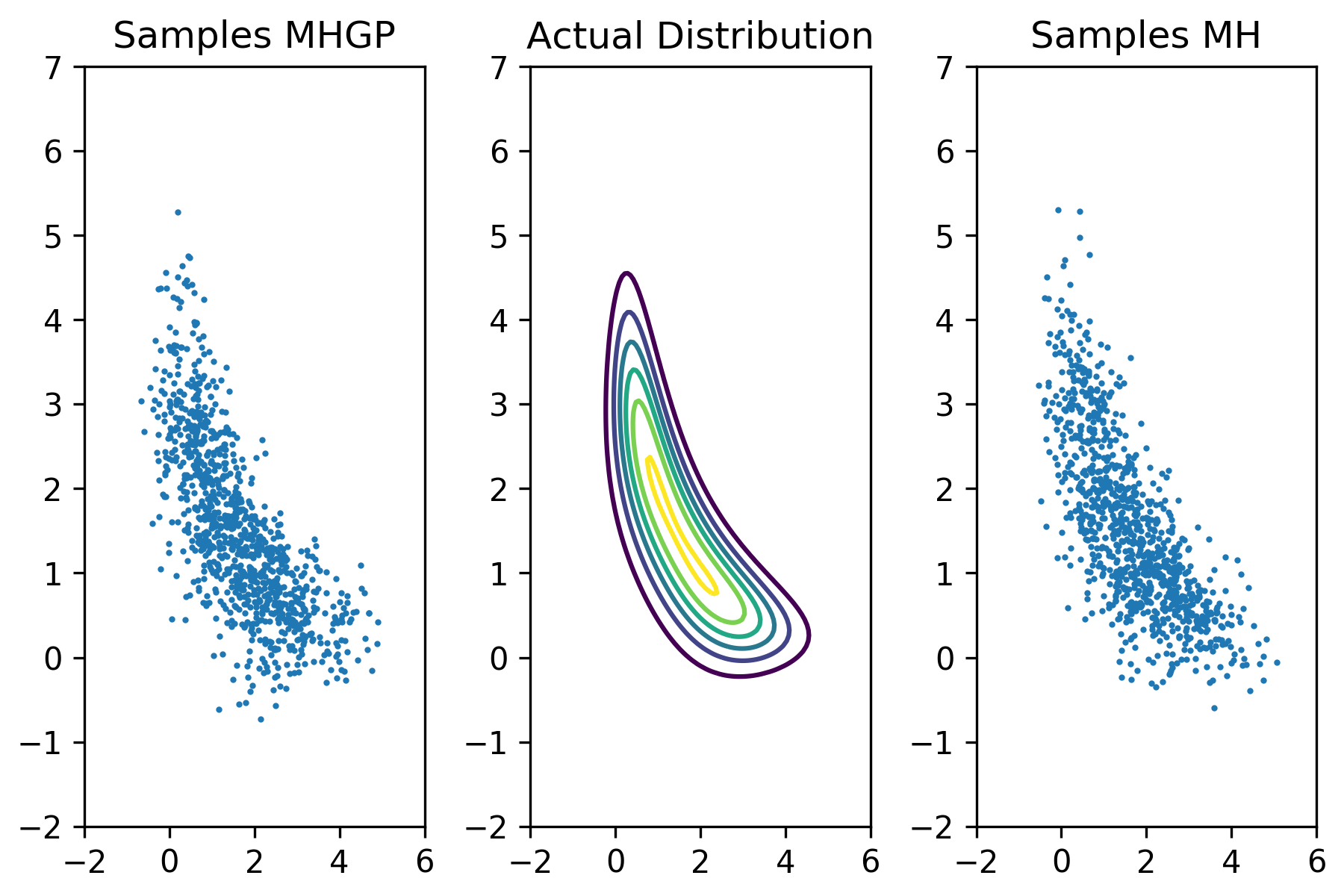}}
\end{minipage}
\caption{Random samples taken from the generated samples for both MHGP and plain MH along with the actual banana distribution in the middle.}
\label{fig:banana}
\end{figure}

For the current point $ x $ and new proposed point $ x^* $ the GP prediction gives us the covariance matrix containing $ \sigma{}_{xx}^2, \sigma{}_{x^*x^*}^2, \sigma{}_{xx^*}^2 $ where $ \sigma{}_{xx}^2 $ and $ \sigma{}_{x^*x^*}^2 $ are the mean-squared errors at points $ x $ and $ x^* $, respectively, and $ \sigma{}_{xx^*}^2 $ is the covariance between $ x $ and $ x^* $. GP is built on the log of the target pdf. Thus, each of the values $\ln{p(x^*|D)}$ and  $\ln{p(x|D)}$ are Gaussian distributed. It makes the log of the acceptance ratio (the ratio between $p(x^*|D)$ and $p(x|D)$) also Gaussian and the acceptance ratio a log-normal random variable, mean of which is used as the measure for the acceptance ratio. The mean formula $\mathrm{e}^{\mu{} + \sigma{}^2/2}$ for log-normal distribution gives $e^{\mu{}^*-\mu{}+\frac{\sigma{}_{xx}^2 + \sigma{}_{x^*x^*}^2 - 2\sigma{}_{xx^*}^2}{2}}$.

Each time a new point $ x^* $ is proposed from the proposal distribution, we measure how certain our GP is about the acceptance probability, $a$, there. The measurement is done by computing $ \sqrt{Var[a]}/E[a] $ and a check is made whether it is larger than some threshold value. Based on the value of the computation being larger than the threshold or not, we decide whether to read the $ p(x^*) $ from the GP or to evaluate the target distribution. To calculate $ \sqrt{Var[a]}/E[a] $ for the log-normal distribution, we use the standard mean and variance formula for log-normal, which give us $ \sqrt{\mathrm{e}^\sigma{}^2 - 1} $ as our desired ratio. Since $ \ln{a} $ was a subtraction of two Gaussian random variables, the value of $ \sigma{}^2 $ will be $ \sigma{}_{xx}^2 + \sigma{}_{x^*x^*}^2 - 2\sigma{}_{xx^*}^2 $.

To limit the time required to train the GP on all the accepted points after each evaluation, we instead used local Gaussian process before making a prediction that considered only the points in the vicinity of the current and the proposed points. We used the well known squared exponential kernel for GP along with automatic relevance determination (ARD). The covariance for the proposal (obtained from Laplace approximation) was scaled down by a configurable parameter.

\section{Experimental Results}

\begin{figure}[thb]

\begin{minipage}[b]{1\linewidth}
  \centering
  \centerline{\includegraphics[width=8cm]{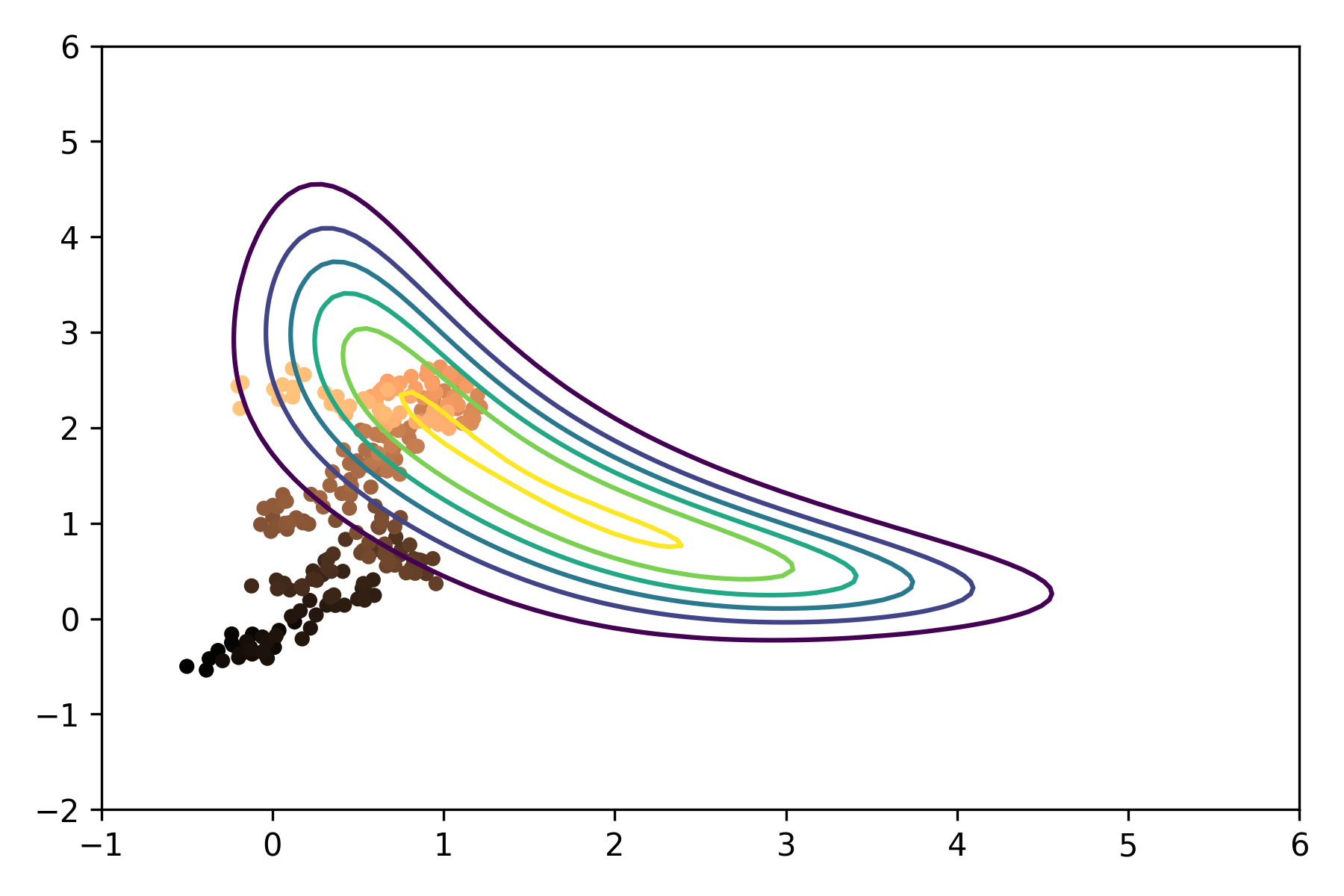}}
\end{minipage}

\begin{minipage}[b]{1\linewidth}
  \centering
  \centerline{\includegraphics[width=8cm]{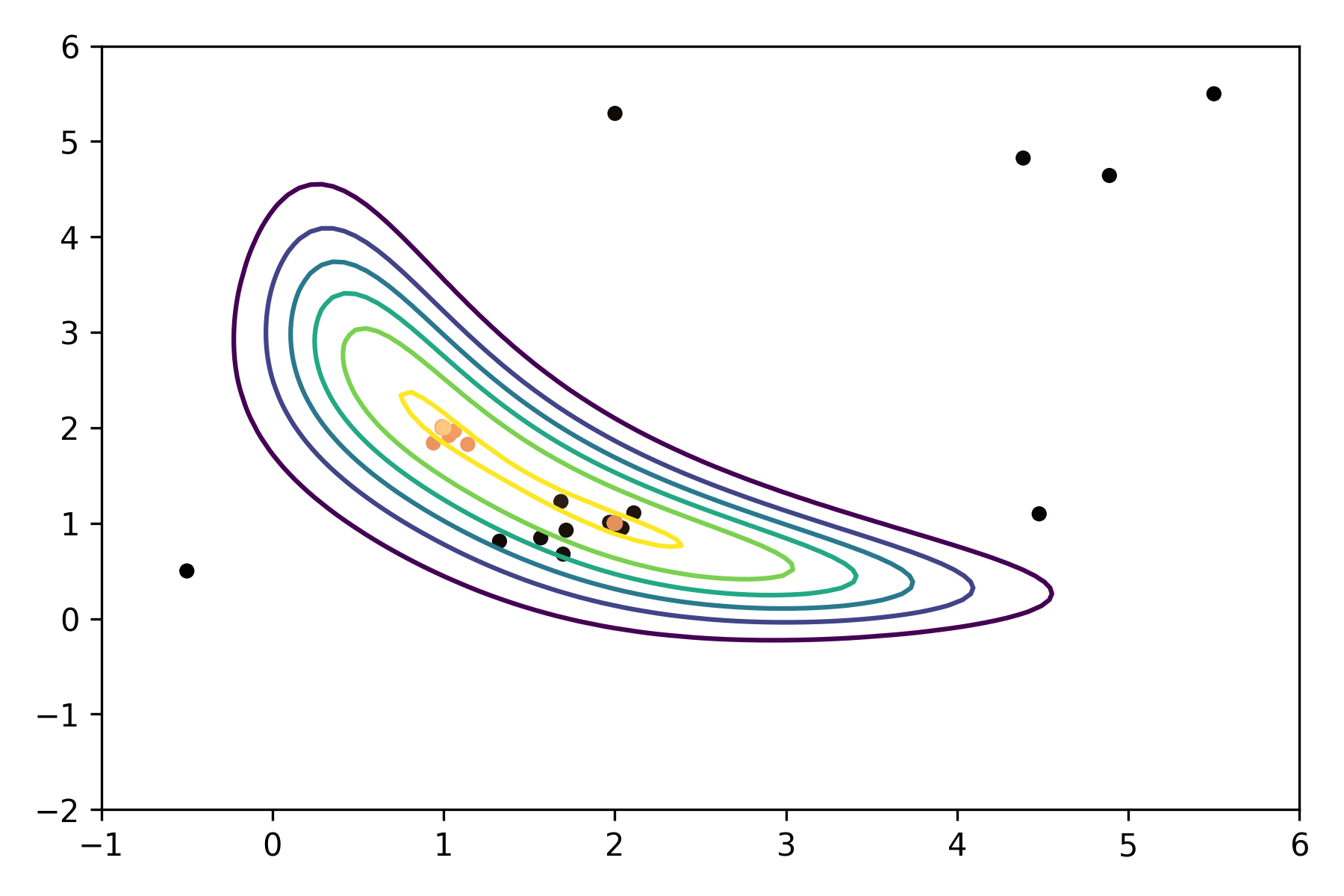}}
\end{minipage}
\caption{Comparison between time taken to reach high density region for plain MH versus the Bayesian optimization phase of MHGP. Top: showing first $250$ iterations of plain MH for sampling banana distribution. Bottom: points explored by Bayesian optimizer for the same problem.}
\label{fig:bo_mh}
\end{figure}

\textbf{Experiment 1.} The comparison between the plain MH and the proposed MHGP was done using a number of experiments, two of which are presented here. The model for the first experiment was the banana distribution~\cite{mcmcdram,banana}. Both the algorithms were run $15000$ iterations to generate samples from the same banana distribution. For plain MH, each iteration needs one target evaluation. MHGP, in contrast, needed less than $200$ evaluations during all these iterations with $50$ additional evaluations during the Bayesian optimization phase. Figure~\ref{fig:banana} shows $500$ random samples generated by both the methods along with the actual distribution and it illustrates that MHGP achieves very similar results as MH but with far less computational cost. The performance gain from using Bayesian Optimization is evident from Figure~\ref{fig:bo_mh}. Both the algorithms started far from the high density region. The plot shows that plain MH required significantly larger number of evaluations compared to the Bayesian optimizer in MHGP.

\begin{figure*}[htb]

\begin{minipage}[b]{0.48\linewidth}
  \centering
  \centerline{\includegraphics[width=10.0cm]{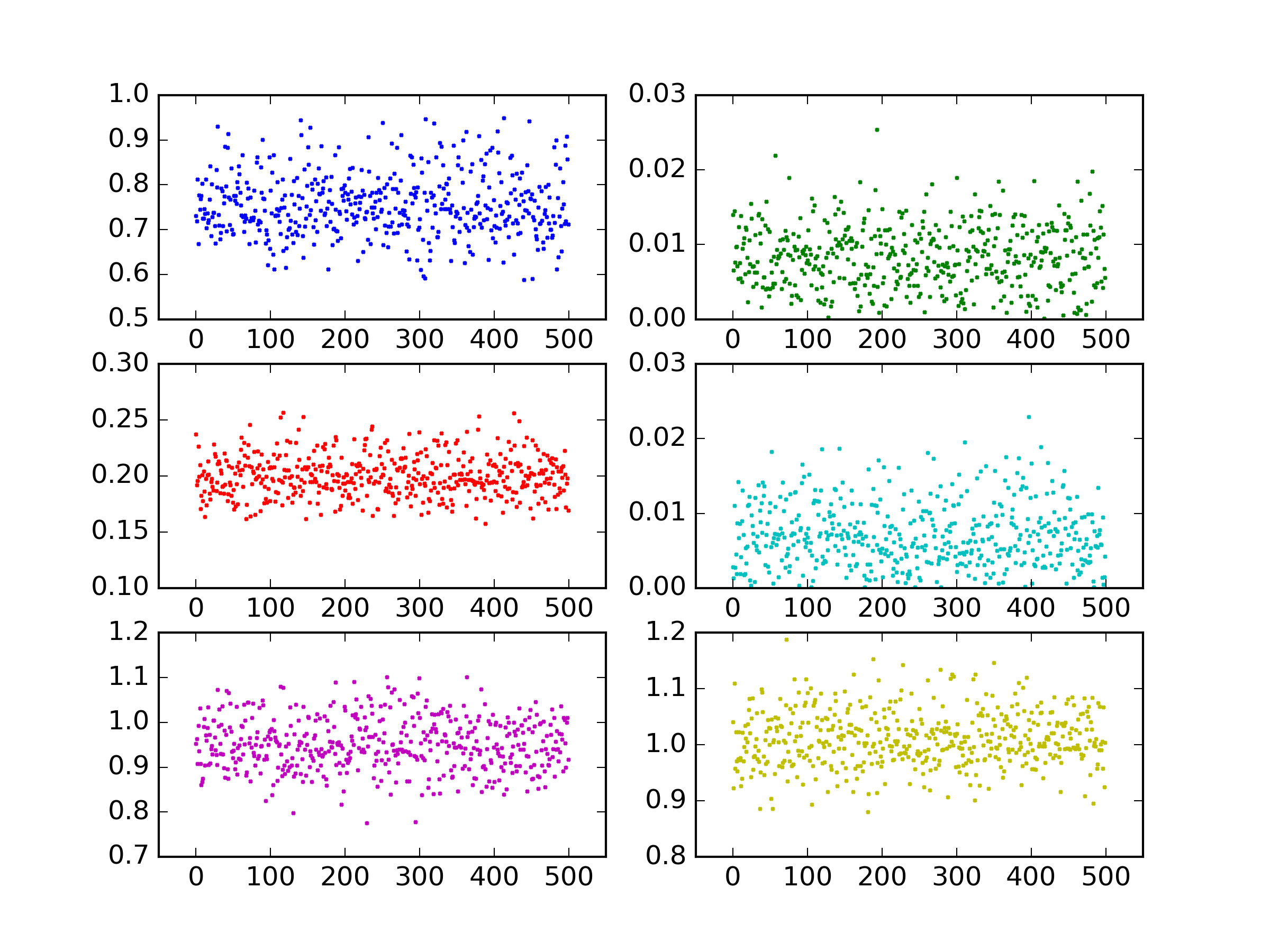}}
\end{minipage}
\hfill
\begin{minipage}[b]{0.48\linewidth}
  \centering
  \centerline{\includegraphics[width=10.0cm]{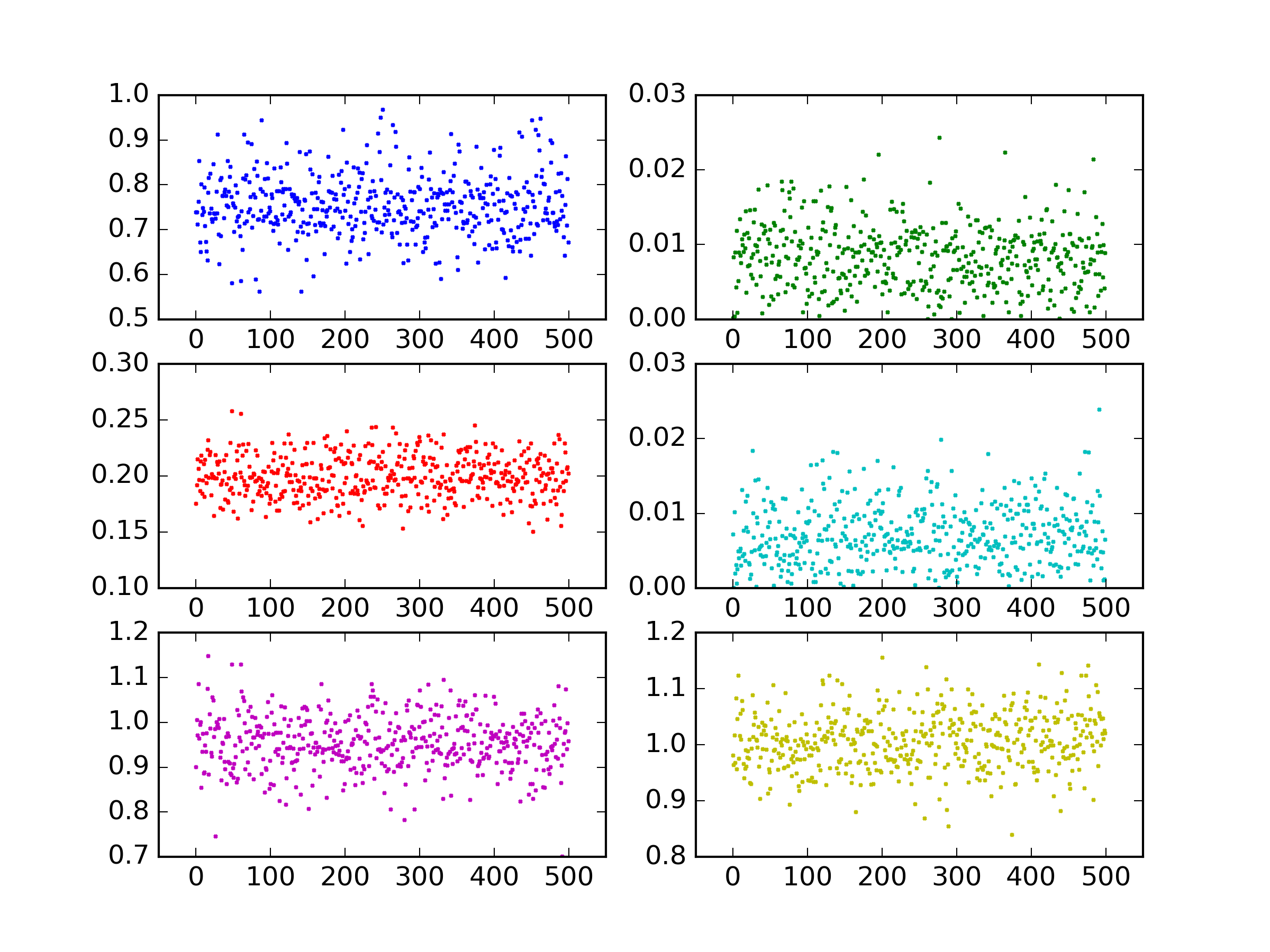}}
\end{minipage}
\caption{Samples along each dimension of the kinetics example. Left: samples generated by MCMC DRAM and Right: samples from MHGP. For both sides, the six plots read from left to right and top to bottom are for the first reaction rate, first activation energy, second reaction rate, second activation energy, and A's initial concentration for the first and second batch, respectively.}
\label{fig:kin}
\end{figure*}

\textbf{Experiment 2.} Model for the second example is a more complex ordinary differential equation to model a chemical kinetics problem. Here, a two step reaction A~$\rightarrow$~B~$\rightarrow$~C was considered with temperature dependent reaction rates~\cite{mcmcdram,boxo}. The dataset consists of two batches of data for two temperature settings where both the batches contain the relative concentrations of A and B over different time steps. There are six unknowns in the model: two reaction rate parameters, two activation energies, and for both the batches - the initial concentration of A. For both MHGP and MCMC DRAM, $500$ random samples out of the accepted samples along each dimension are shown in Figure~\ref{fig:kin}. 

\begin{figure}[bht]
\begin{minipage}[b]{1\linewidth}
  \centering
  \centerline{\includegraphics[width=8.5cm]{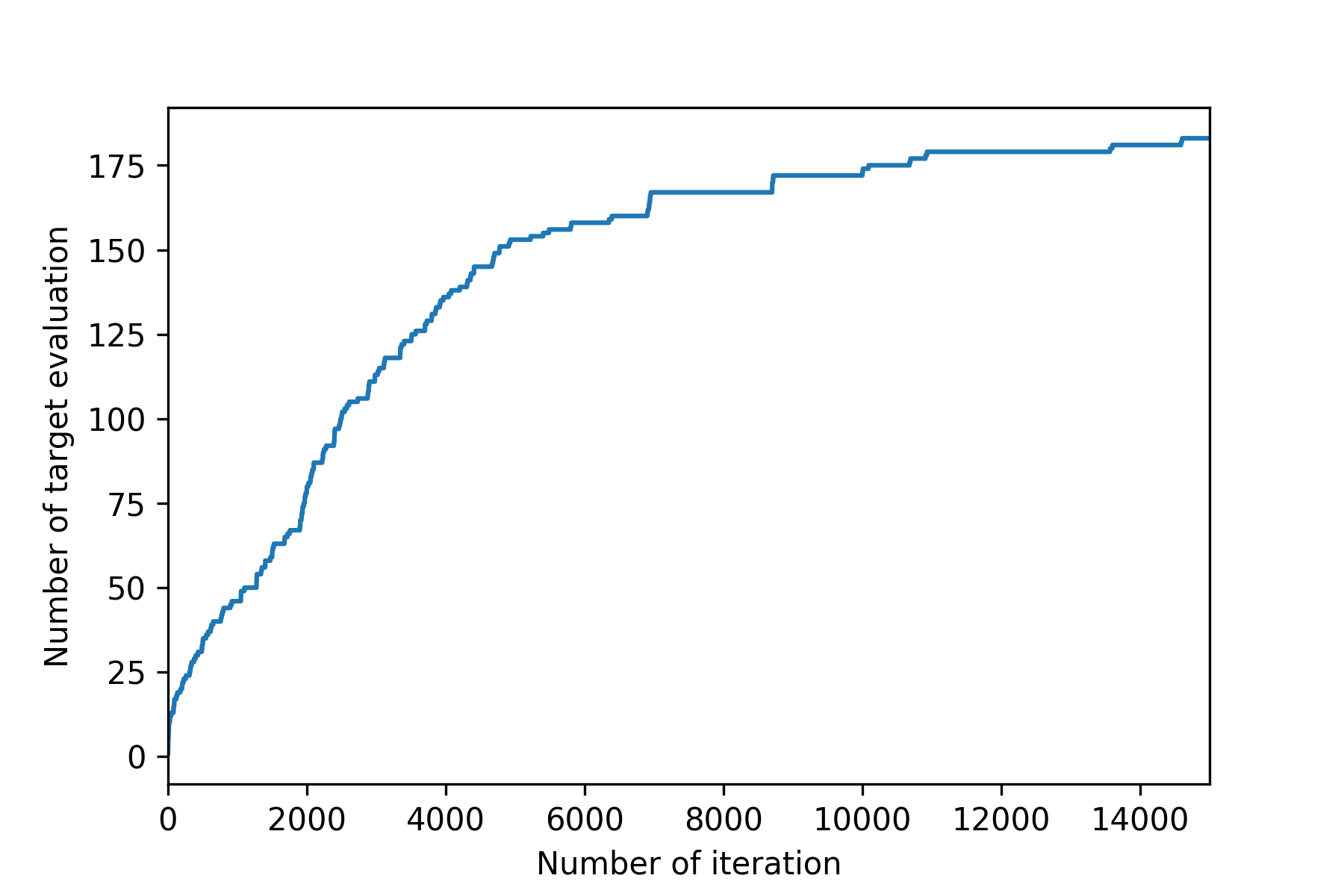}}
\end{minipage}
\caption{Target evaluations of MHGP.}
\label{fig:train}
\end{figure}

The samples from MHGP again has similar distribution to that of the established method. We performed statistical test based on the energy distance ~\cite{energystatistic} measures between the two sets of samples generated by the two methods to find if the samples indeed come from the same distribution. No statistically significant difference was found between the two sets with p-value of $0.12$ for the kinetics example and $0.15$ for the banana distribution. The fact that MHGP, driven by the uncertainty measurement from Gaussian process, requires less and less target evaluations as the algorithm advances through the iterations, can be observed in Figure~\ref{fig:train}. GP starts with high uncertainty and many of the initially proposed points need to be evaluated. But gradually it gains a better approximation of the target and very few evaluations are needed in later stages.

\section{Conclusion}

The key challenge in this work was to reduce the number of costly evaluations while ensuring efficient convergence to the target distribution. As our experiments and corresponding comparative study have indicated, MHGP offers an efficient alternative to the plain Metropolis-Hastings. It has short burn-in period with the help of Bayesian optimization, an informed proposal distribution using Laplace approximation, and fewer target evaluations due to Gaussian process with quantified uncertainty. It, however, suffers from the same limitations that the MCMC methods typically do, such as lack of support for multi-modal distributions. Also, since the method is based on a GP approximation of the target, adherence to the detailed balance property cannot be established. Nevertheless, we believe the method can have significant practical value for different areas of science and engineering where forward simulations are expensive.

\section{Acknowledgment}

This material is based upon work supported by the National Institute of Food and Agriculture (NIFA)/USDA under Grand No. 2017-67017-26167 and the National Science Foundation under Grant No. DMREF-1534260.

\vfill\pagebreak

\bibliographystyle{IEEEbib}
\bibliography{MHGP}

\begin{thebibliography}{10}

\bibitem{a1mcmc}
Christophe Andrieu, Nando~De Freitas, Arnaud Doucet, and Michael~I. Jordan,
\newblock ``{An Introduction to MCMC for Machine Learning},''
\newblock {\em Machine Learning}, vol. 50, pp. 5--43, 2003.

\bibitem{prevwork2}
Jonathan M.~R. Byrd, Stephen~A. Jarvis, and Abhir~H. Bhalerao,
\newblock ``{Reducing the run-time of MCMC programs by multithreading on SMP
  architectures},''
\newblock in {\em Parallel and Distributed Processing, 2008. IPDPS 2008. IEEE
  International Symposium on}, 2008, pp. 1--8.

\bibitem{prevwork3}
Sungjin Ahn, Babak Shahbaba, and Max Welling,
\newblock ``{Distributed Stochastic Gradient MCMC},''
\newblock in {\em Proceedings of the 31st International Conference on Machine
  Learning (ICML-14)}. 2014, pp. 1044--1052, JMLR Workshop and Conference
  Proceedings.

\bibitem{prevwork1}
Anoop Korattikara, Yutian Chen, and Max Welling,
\newblock ``{Austerity in MCMC Land: Cutting the Metropolis-Hastings Budget},''
  2013.

\bibitem{prevwork5}
Matias Quiroz, Mattias Villani, and Robert Kohn,
\newblock ``{Speeding Up MCMC by Efficient Data Subsampling},'' 2016.

\bibitem{gphmc}
Carl~Edward Rasmussen,
\newblock ``{Gaussian Processes to Speed up Hybrid Monte Carlo for Expensive
  Bayesian Integrals},''
\newblock in {\em Bayesian Statistics 7}, J.~M. Bernardo, M.~J. Bayarri, J.~O.
  Berger, A.~P. Dawid, D.~Heckerman, A.~F.~M. Smith, and M.~West, Eds., pp.
  651--659. Oxford University Press, 2003.

\bibitem{simwork}
Colin~Fox J.~Andrés~Christen,
\newblock ``{Markov Chain Monte Carlo Using an Approximation},''
\newblock {\em Journal of Computational and Graphical Statistics}, vol. 14, no.
  4, pp. 795--810, 2005.

\bibitem{ownwork}
Asif~J. Chowdhury and Gabriel~A. Terejanu,
\newblock ``{An Enhanced Metropolis-Hastings Algorithm Based on Gaussian
  Processes},''
\newblock in {\em IMAC XXXIV}, 2016.

\bibitem{Haario99adaptiveproposal}
Heikki Haario, Eero Saksman, and Johanna Tamminen,
\newblock ``{Adaptive proposal distribution for random walk Metropolis
  algorithm},'' Sep 1999.

\bibitem{mcmcdram}
Heikki Haario, Marko Laine, Antonietta Mira, and Eero Saksman,
\newblock ``{DRAM: Efficient adaptive MCMC},''
\newblock {\em Statistics and Computing}, vol. 16, no. 4, pp. 339--354, Dec
  2006.

\bibitem{Larjo2015}
Antti Larjo and Harri L{\"a}hdesm{\"a}ki,
\newblock ``{Using multi-step proposal distribution for improved MCMC
  convergence in Bayesian network structure learning},''
\newblock {\em EURASIP Journal on Bioinformatics and Systems Biology}, vol.
  2015, no. 1, pp. 1--14, 2015.

\bibitem{NIPS2014_5594}
Franziska Meier, Philipp Hennig, and Stefan Schaal,
\newblock ``{Incremental Local Gaussian Regression},''
\newblock in {\em Advances in Neural Information Processing Systems 27},
  Z.~Ghahramani, M.~Welling, C.~Cortes, N.~D. Lawrence, and K.~Q. Weinberger,
  Eds., pp. 972--980. Curran Associates, Inc., 2014.

\bibitem{lgp}
D.~Nguyen-Tuong, M.~Seeger, and J.~Peters,
\newblock ``{Local Gaussian Process Regression for Real Time Online Model
  Learning and Control},''
\newblock in {\em Advances in neural information processing systems 21}, Red
  Hook, NY, USA, June 2009, Max-Planck-Gesellschaft, pp. 1193--1200, Curran.

\bibitem{Shahriari:2015}
Bobak Shahriari, Kevin Swersky, Ziyu Wang, Ryan~P. Adams, and Nando de~Freitas,
\newblock ``{Taking the Human Out of the Loop: A Review of Bayesian
  Optimization},''
\newblock Tech. {R}ep., Universities of Harvard, Oxford, Toronto, and Google
  DeepMind, 2015.

\bibitem{banana}
``{Banana example},''
  \url{https://mjlaine.github.io/mcmcstat/ex/bananaex.html},
\newblock Accessed: 2019-10-21.

\bibitem{boxo}
``{Boxo chemical kinetics example},''
  \url{https://mjlaine.github.io/mcmcstat/ex/boxoex.html},
\newblock Accessed: 2019-10-21.

\bibitem{energystatistic}
Gábor~J. Székely and Maria~L. Rizzo,
\newblock ``{Energy statistics: A class of statistics based on distances},''
\newblock {\em Journal of Statistical Planning and Inference}, vol. 143, no. 8,
  pp. 1249 -- 1272, 2013.

\end{thebibliography}

\end{document}